\documentclass[10pt,twocolumn,letterpaper]{article}
\pdfoutput=1
\usepackage{iccv}
\usepackage{times}
\usepackage{epsfig}
\usepackage{graphicx}
\usepackage{amsmath}
\usepackage{amssymb}

\usepackage[breaklinks=true,bookmarks=false]{hyperref}

\iccvfinalcopy 

\usepackage{booktabs}
\usepackage{multirow}
\usepackage{color}
\usepackage{algorithm}
\usepackage{algorithmic}
\usepackage{float}
\usepackage[table,xcdraw]{xcolor}

\def\iccvPaperID{****} 

\ificcvfinal\fi

\begin{document}

\title{Domain Generalization via Balancing Training Difficulty and Model Capability}

\author{Xueying Jiang, Jiaxing Huang, Sheng Jin, Shijian Lu\thanks{Corresponding author.} \\
S-Lab, Nanyang Technological University\\
{\tt\small xueying003@e.ntu.edu.sg}\\
{\tt\small \{Jiaxing.Huang, Sheng.Jin, Shijian.Lu\}@ntu.edu.sg}
}

\maketitle
\ificcvfinal\thispagestyle{empty}\fi

\begin{abstract}

Domain generalization (DG) aims to learn domain-generalizable models from one or multiple source domains that can perform well in unseen target domains. Despite its recent progress, most existing work suffers from the misalignment between the difficulty level of training samples and the capability of contemporarily trained models, leading to over-fitting or under-fitting in the trained generalization model. We design MoDify, a Momentum Difficulty framework that tackles the misalignment by balancing the seesaw between the model's capability and the samples' difficulties along the training process. MoDify consists of two novel designs that collaborate to fight against the misalignment while learning domain-generalizable models. The first is MoDify-based Data Augmentation which exploits an RGB Shuffle technique to generate difficulty-aware training samples on the fly. The second is MoDify-based Network Optimization which dynamically schedules the training samples for balanced and smooth learning with appropriate difficulty. Without bells and whistles, a simple implementation of MoDify achieves superior performance across multiple benchmarks. In addition, MoDify can complement existing methods as a plug-in, and it is generic and can work for different visual recognition tasks.

\end{abstract}
\section{Introduction}

\begin{figure}
\centering
\includegraphics[width=1\linewidth]{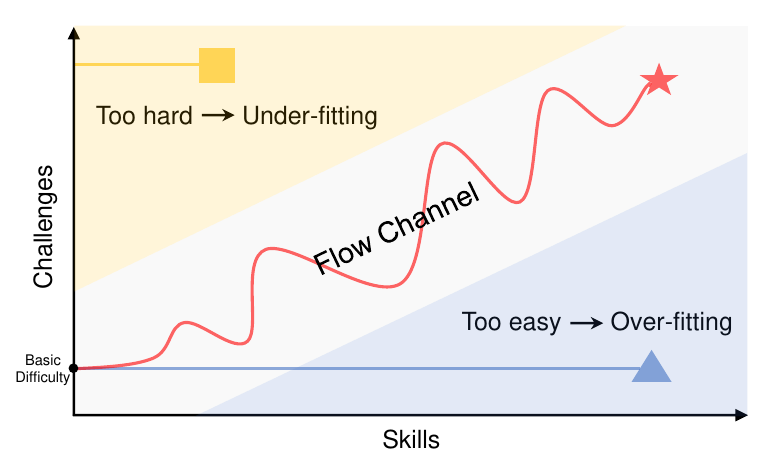}
\caption{
Illustration of the proposed MoDify framework.
Training domain-generalizable models often suffer from clear under-fitting (or over-fitting) if keep feeding over-difficult (or over-easy) training samples, especially at the early (or later) training stage, both leading to degraded generalization of the trained models (as illustrated in \textcolor[RGB]{238,173,14}{yellow}/\textcolor[RGB]{65,105,225}{blue} lines).
Inspired by the Flow Theory~\cite{csikszentmihalyi1990flow} that a learner usually has better learning outcome when the learner's skill and the task difficulty are well aligned (i.e., lying within the \textit{Flow Channel}), the proposed MoDify schedules the training samples adaptively according to the alignment between the sample difficulty and the capability of contemporarily trained models (as illustrated in \textcolor[RGB]{255,0,0}{red} line).
}
\label{fig:flow_theory}
\end{figure}

Deep neural networks (DNNs)~\cite{he2016deep, krizhevsky2017imagenet, simonyan2014very} have achieved significant progress in recent years with numerous network architectures and learning algorithms designed for various discriminative tasks. In the area of computer vision, DNNs have achieved great success in various visual recognition tasks such as image segmentation~\cite{chen2017deeplab, zhao2017pyramid, xie2021segformer}, object detection~\cite{ren2015faster, carion2020end}, etc.
However, deep network training often suffers from a misfitting problem, being either over-fitting or under-fitting due to the misalignment between the capacity of networks under training and the complexity of training data.
While applying a misfit deep network model to the data from a different domain, the misfitting problem can be greatly enlarged due to the distribution bias and distribution shift across domains.


Domain generalization aims to mitigate the misfitting problem by learning a domain generalizable model that can work well in new domains. It has been widely studied via different augmentation strategies, e.g., domain randomization~\cite{peng2021global, yue2019domain, huang2021fsdr, wu2022siamdoge}, feature augmentation~\cite{lee2022wildnet, pan2019switchable, choi2021robustnet}, and data augmentation~\cite{zhao2022style, pan2018two}, targeting to obtain generalization capability by \textit{seeing} more training data with various diverse characteristics. However, the aforementioned methods mostly neglect the misalignment between the difficulty level of training samples and the capability of the contemporary models along the training process,  leading to misfit deep network models and degraded performance.

The Flow Theory~\cite{csikszentmihalyi1990flow} has been widely studied in the field of learning and education, which suggests that a learner has optimal learning outcomes when the capability of the learner is well aligned with the difficulty level of learning tasks throughout the learning process. Inspired by this theory, we design \textbf{MoDify}, a \textbf{Mo}mentum \textbf{Dif}ficult\textbf{y} framework that aims to tackle the misfitting problem in deep network training. The idea is to dynamically gauge the difficulty level of training samples along the training process, and feed training samples whose difficulty level is well aligned with the capability of the contemporary deep network model under training. This directly leads to a balanced learning process between the difficulty level of training samples and the model capability as illustrated in Fig.~\ref{fig:flow_theory}, which helps mitigate the misfitting problem effectively.


MoDify consists of two novel designs for balanced and smooth learning. The first is MoDify-based \textbf{D}ata \textbf{A}ugmentation (\textbf{MoDify-DA}) that produces augmented training samples with relevant difficulty levels on the fly. The second is MoDify-based \textbf{N}etwork \textbf{O}ptimization (\textbf{MoDify-NO}) that achieves progressive network training by considering the difficulty level of training samples.
The two designs work in a collaborative manner to maintain the difficulty-capability balance, which coordinate the augmentation and network training smoothly according to the model's capability.
Moreover, we employ an efficient yet effective RGB Shuffle technique that enables online sample augmentation by shuffling the color channels while preserving spatial structures efficiently. RGB Shuffle improves the generalization of the trained model effectively. MoDify has three desirable features: 1) it is generic and performs well across different visual recognition tasks such as image semantic segmentation and object detection; 2) it is an online technique with negligible computational cost; 3) it is complementary with existing DG methods and can be incorporated with consistent performance boosts.


In summary, the contributions of this work are threefold.
\textit{First}, we propose MoDify, a novel momentum difficulty framework that effectively addresses the network misfitting problem by maintaining the balance between the difficulty level of training samples and the capability of the contemporary models along the training process.
\textit{Second}, we design MoDify-DA and MoDify-NO, the former generates difficulty-aware augmentation samples on the fly while the latter coordinates for a smooth learning process by dropping over-simple samples and postponing over-difficult samples to a later training phase.
\textit{Third}, extensive experiments show that a simple implementation of MoDify achieves superior performance consistently across multiple benchmarks and visual recognition tasks.



\section{Related Work}

\textbf{Domain generalization (DG)} aims to generalize the model learned on one or multiple domains to unseen target domains which have been explored in various computer vision tasks, such as object detection~\cite{pan2018two, huang2021fsdr, lin2021domain}, semantic segmentation~\cite{pan2018two, yue2019domain, peng2021global, huang2021fsdr, lee2022wildnet, zhao2022style}. Most existing DG methods can be broadly categorized into single-source DG~\cite{huang2021fsdr, choi2021robustnet, pan2018two, yue2019domain, peng2021global, zhao2022style, wu2022siamdoge, pan2019switchable, lee2022wildnet, zhou2020learning} and multi-source DG~\cite{zhao2020domain, motiian2017unified, matsuura2020domain, jia2020single, balaji2018metareg, li2019episodic, du2020metanorm, du2020learning, dou2019domain, choi2021robustnet, zhao2022style, qiao2020learning, zhao2020maximum}, both targeting to learn domain-invariant feature representations from various aspects, including \textit{domain alignment}~\cite{zhao2020domain, motiian2017unified, matsuura2020domain, jia2020single}, \textit{meta-learning}~\cite{balaji2018metareg, li2019episodic, du2020metanorm, du2020learning, dou2019domain} and \textit{augmentation strategies}~\cite{huang2021fsdr, choi2021robustnet, pan2018two, yue2019domain, peng2021global, zhao2022style, wu2022siamdoge, pan2019switchable, lee2022wildnet}.
Our work belongs to single-source DG, aiming to
address a more challenging issue when only one single source domain is available during training.

Single-source DG usually works by domain randomization that augments data~\cite{zhao2022style, zhou2020learning, qiao2020learning, zhao2020maximum} or domains~\cite{peng2021global, yue2019domain, huang2021fsdr, wu2022siamdoge}. Most existing methods aim to enhance the variation of synthetic images in a source domain by adversarial data augmentation~\cite{qiao2020learning, zhao2020maximum} or designing customized modules~\cite{zhao2022style, zhou2020learning}. However, they largely neglect the misalignment between the difficulty level of training samples and the capability of contemporary models during training, leading to degraded generalization performance. In this work, we design an effective and efficient strategy to address the misfitting problem.

\begin{figure*}[htbp]
	\centering
	\includegraphics[width=1\linewidth]{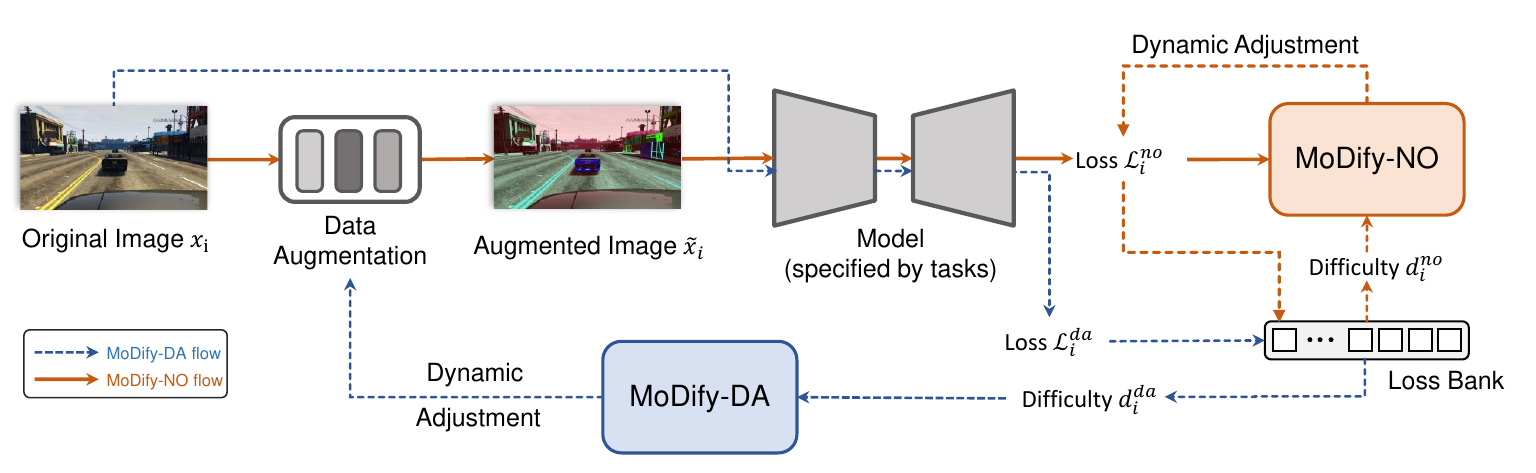}

	\caption{Overall architecture of the proposed Momentum Difficulty (MoDify). In the MoDify-DA flow (highlighted by \textcolor[RGB]{65,105,225}{blue arrows}), the network takes the original image $x_i$ as input and generates its loss $\mathcal{L}^{da}_i$, and applies the loss to compute the difficulty level $d^{da}_i$ with the Loss Bank. MoDify-DA dynamically adjusts the strength of data augmentation according to the $d^{da}_i$. In the MoDify-NO flow (highlighted by \textcolor[RGB]{255,0,0}{red arrows}), the network takes the augmented image $\widetilde{x}_i$ as input (with a difficulty level of $d^{da}_i$). Then the difficulty degree $d^{no}_i$ of the augmented image $\widetilde{x}_i$ is calculated in the same way. MoDify-NO decides whether postpone, drop, or learn from this sample based on the $d^{no}_i$ from the Loss Bank. Noted the sample is fed for training only if its difficulty level is aligned with the model's capability. Additionally, MoDify-DA introduces little computational overhead without involving any back propagation. 
 }
	\label{fig:overall_architecture}
\end{figure*}

\textbf{Flow Theory}~\cite{csikszentmihalyi2013flow, csikszentmihalyi1997flow, csikszentmihalyi1990flow}, which is a well-established theory in psychology, suggests that optimal learning could be achieved when the skills level of the learner is aligned with the difficulty level of the tasks during learning. It has been extensively studied in the field of education~\cite{csikszentmihalyi1997education, heutte2016proposal} and has more recently been applied to game designs~\cite{chen2007flow, kiili2012design}. We introduce the Flow Theory into computer vision research for tackling the domain generalization challenge. The idea is that domain generalization often suffers from unbalance between the difficulty level of training samples and the capability of contemporarily trained models. Flow theory can fit in perfectly by scheduling the training samples according to their difficulties while training domain generalizable networks.

\textbf{Curriculum Learning}~\cite{bengio2009curriculum} is a widely studied learning strategy, which involves starting with easier training samples~\cite{wei2016stc, chen2015webly, guo2018curriculumnet} or sub-tasks~\cite{ao2021co, pentina2015curriculum, narvekar2016source, matiisen2019teacher} and gradually increasing the difficulty level. It has attracted increasing attention recently with the advance of deep learning, and it has been studied in various visual recognition tasks such as domain adaption~\cite{zhang2019curriculum, choi2019pseudo}.
For instance, ~\cite{zhang2019curriculum} divides the semantic segmentation task into sub-tasks and learns from easy to difficult ones, aiming to decrease the learning difficulty in the early training period.
~\cite{choi2019pseudo} ranks the training samples according to pseudo-label correctness probabilities and learns from them sequentially during training.
However, most existing curriculum learning methods require the pre-defined difficulty levels of the samples in training. We instead dynamically augment and feed difficulty-aware training samples according to the capability of contemporarily trained network models along the training process.


\section{Method}
This section presents the proposed \textbf{Mo}mentum \textbf{Dif}ficult\textbf{y} (\textbf{MoDify}) framework. First, the problem definition and overview are presented in Sec.~\ref{subsec:problem_definition} and Sec.~\ref{subsec:overview}, respectively. The detailed designs of MoDify are then introduced, including Loss Bank-based Difficulty Assessment in Sec.~\ref{subsec:loss_bank} and Difficulty-Aware Training Strategy in Sec.~\ref{subsec:difficulty_aware_training_strategy}.
Finally, loss functions are presented in Sec.~\ref{subsec:loss}.

\subsection{Problem Definition \label{subsec:problem_definition}}

Domain generalization (DG) aims to learn a generalizable model (trained in source domain $\mathcal{S}$) that can work in various unseen target domains $\mathcal{T} = \left \{ \mathcal{T}_{1}, ..., \mathcal{T}_{K}  \right \} $. During training, only the dataset $D^{s} =\left \{ \left ( x^{s}, y^{s}   \right )  \right \} $ of the source domain is available.


\subsection{Momentum Difficulty (MoDify)\label{subsec:overview}}
The proposed MoDify aims to address the imbalance issue between the difficulty level of augmented training samples and the capability of contemporarily trained models while training domain generalizable networks. It tackles this challenge by augmenting and scheduling difficulty-aware training samples dynamically according to the capability of contemporarily trained network models.




\textbf{Overview.}
Fig.~\ref{fig:overall_architecture} illustrates the overall architecture of the proposed MoDify, which includes a loss bank and a difficulty-aware training framework comprising
two specific strategies (i.e. MoDify-DA and MoDify-NO). 

In the MoDify-DA flow, the original image $x_i$ is fed into the visual task network to obtain its loss $\mathcal{L}^{da}_{i}$. Then, the loss bank takes $\mathcal{L}^{da}_{i}$ as input and outputs the difficulty degree $d^{da}_i$ of $x_i$, where the loss bank is updated in a momentum-based manner. Finally, the augmented samples $\widetilde{x}_i$ are generated based on the difficulty degree $d^{da}_i$.
In the MoDify-NO flow, the difficulty degree $d^{no}_i$ of the augmented image $\widetilde{x}_i$ is obtained using the same approach as in the MoDify-DA flow. The network is updated only when the value of $d^{no}_i$ falls within a moderate range, allowing it to prioritize learning samples with appropriate difficulty.

This dual flow mechanism ensures that the model learns samples whose difficulty is aligned with the model's capability, thus making the training process more efficient and smooth by rejecting undesirably samples.

\subsection{Loss Bank-based Difficulty Assessment \label{subsec:loss_bank}}
The Loss Bank is a crucial component of MoDify that establishes a consistent measure of training samples' difficulty by maintaining a list containing loss values for each sample processed by the visual task network.
The size of the Loss Bank is based on the total number of samples in the training dataset, rather than the mini-batch size, providing a global-scale measurement of the training samples' difficulty.
\begin{algorithm}
	\renewcommand{\algorithmicrequire}{\textbf{Input:}}
	\renewcommand{\algorithmicensure}{\textbf{Output:}}
	\caption{Loss Bank during training}
	\label{alg1:loss_bank}
	\begin{algorithmic}[1]
		\STATE Initialization:
  \STATE$B = \{ V_{i} = \alpha \mid i \in \{1, ..., N \} \}$, epoch $j$
            \FOR{$j=1$ \textbf{to} $j=M$}
                \FOR{$i=1$ \textbf{to} $i=N$}

                \STATE Update $ V_{i} $ by Eqn.~\ref{eqn_1} and $\mathcal{L}^{da}_{i}$
                \STATE Update $d^{da}_{i}$ by Eqn.~\ref{eqn_dif} and $\mathcal{L}^{da}_{i}$
                \STATE Update $d^{no}_{i}$ by Eqn.~\ref{eqn_dif} and $\mathcal{L}^{no}_{i}$

		      \ENDFOR
            \ENDFOR

	\end{algorithmic}
\end{algorithm}

The overall process of Loss Bank updating is shown in Algorithm~\ref{alg1:loss_bank}. Specifically, the values of Loss Bank $B = \{ V_{i} \mid  i \in \{1, ..., N \} \}$ are defined on-the-fly by a set of data samples. $N$ equals the number of samples in the training dataset.
We adopt a momentum manner to update the Loss Bank during training:
\begin{equation}
    V_{i} = \lambda V^{\prime}_{i} + (1 - \lambda)\mathcal{L}^{da}_{i},
    \label{eqn_1}
\end{equation}

\noindent where $V^{\prime}_{i}$ and $V_{i}$ denote the i-th sample's value of the last epoch and the current epoch separately, and $\lambda$ is the momentum coefficient.

\textbf {Difficulty Degree.} The proposed loss bank is utilized to assess the difficulty level of the samples, which provides a global and dynamic perspective on the samples' loss values. For each sample $x_i$, we first fed it into the visual task network and output its loss $\mathcal{L}_i$, where the loss has a different formulation according to the tasks. Then we use the relative rank of the loss $\mathcal{L}_i$ in the loss bank as the difficulty degree, which is formulated as below:

\begin{equation}
    d_{i} =  \frac{\sum_{k=1}^N I(\mathcal{L}_{i} < V_{k} )}{N}
    \label{eqn_dif},
\end{equation}
where $V_{k} \in B$ represents the loss value of $x_k$ in Loss Bank and $I(x)$ is an indicator function.

\textbf{Remark 1.} \textit{MoDify is an efficient training framework using the lightweight and simple Loss Bank. For instance, in comparison to DG methods using image translation GANs~\cite{rahman2019multi}, which typically utilizes 9 convolutional layers with about 11,000,000 parameters, MoDify is much more efficient, which only utilizes a fixed-length list with approximately $N$ parameters. Here $N$ equals to the size of samples contained in the training dataset.}

\subsection{Difficulty-Aware Training Strategy\label{subsec:difficulty_aware_training_strategy}}

This subsection introduces the designed difficulty-aware training strategies of MoDify, including MoDify-DA and MoDify-NO. Besides, the data augmentation strategy used in MoDify-DA is also introduced.



\textbf {MoDify-DA.} We propose the MoDify-DA strategy that dynamically adjusts the strength of data augmentation. For each input original image $x_i$, MoDify-DA calculates its augmentation degree based on the sample's difficulty degree $d^{da}_i$ using Eqn.~\ref{eqn_dif}. We utilize $1-d^{da}_i$ as the augmentation degree of $x_i$ and use it as the probability to augment the input image so that samples with higher difficulty levels remain unchanged and at the same time simpler samples are augmented.




A simple yet effective data augmentation method is utilized to improve the domain invariance in this section. In Domain Generalization (DG) tasks, learning domain-invariant features is crucial for better generalization performance, especially as the source and target domains frequently differ in style and color but share spatial layout similarities. Leveraging spatial information such as edges and shapes can be beneficial. For instance, while the color of a simulated car and a real car may differ, their shape is often similar. Motivated by this observation, we select an appropriate data augmentation method called RGB Shuffle. This method randomly permutes the R, G, and B channels of a training image, effectively altering its style while preserving its structural information.

Compared with offline data augmentation methods~\cite{ho2019population, cubuk2019autoaugment}, MoDify-DA is designed to perform strategy online. In contrast to online data augmentation techniques like~\cite{xu2022universal}, which involves multi-round perturbation, and~\cite{tang2020onlineaugment}, requiring an extra model for parameter selection of data augmentation, MoDify-DA requires just one additional round for deciding the degree of data augmentation, eliminating the need for an auxiliary parameter optimization model. Therefore, MoDify-DA exhibits more efficiency that brings only little computational cost.


\textbf {MoDify-NO.} We propose MoDify-NO strategy that enables the network focus on samples with a moderate difficulty degree. For each input image $\widetilde{x}_i$,
MoDify-NO decides whether or not to learn from this sample based on its difficulty level $d^{no}_i$ using Eqn.~\ref{eqn_dif}. To achieve this, we dynamically adjust the weight $w_i$ used for the sample's loss function $w_i\mathcal{L}_i$, which is formulated as:


\begin{equation}
w_i= \begin{cases}
1.0,\quad &d^{no}_{i} \in (T_{easy}, T_{hard}) \\
0.0,\quad &others
\end{cases} ,
\end{equation}
where $T_{hard}$ and $T_{easy}$ represent the thresholds for filtering out samples that are either too easy or too difficult. We set $T_{hard}=0.95$ and $T_{easy}=0.05$ in experiments.

\begin{figure}[t]
	\centering
        \includegraphics[width=1\linewidth]{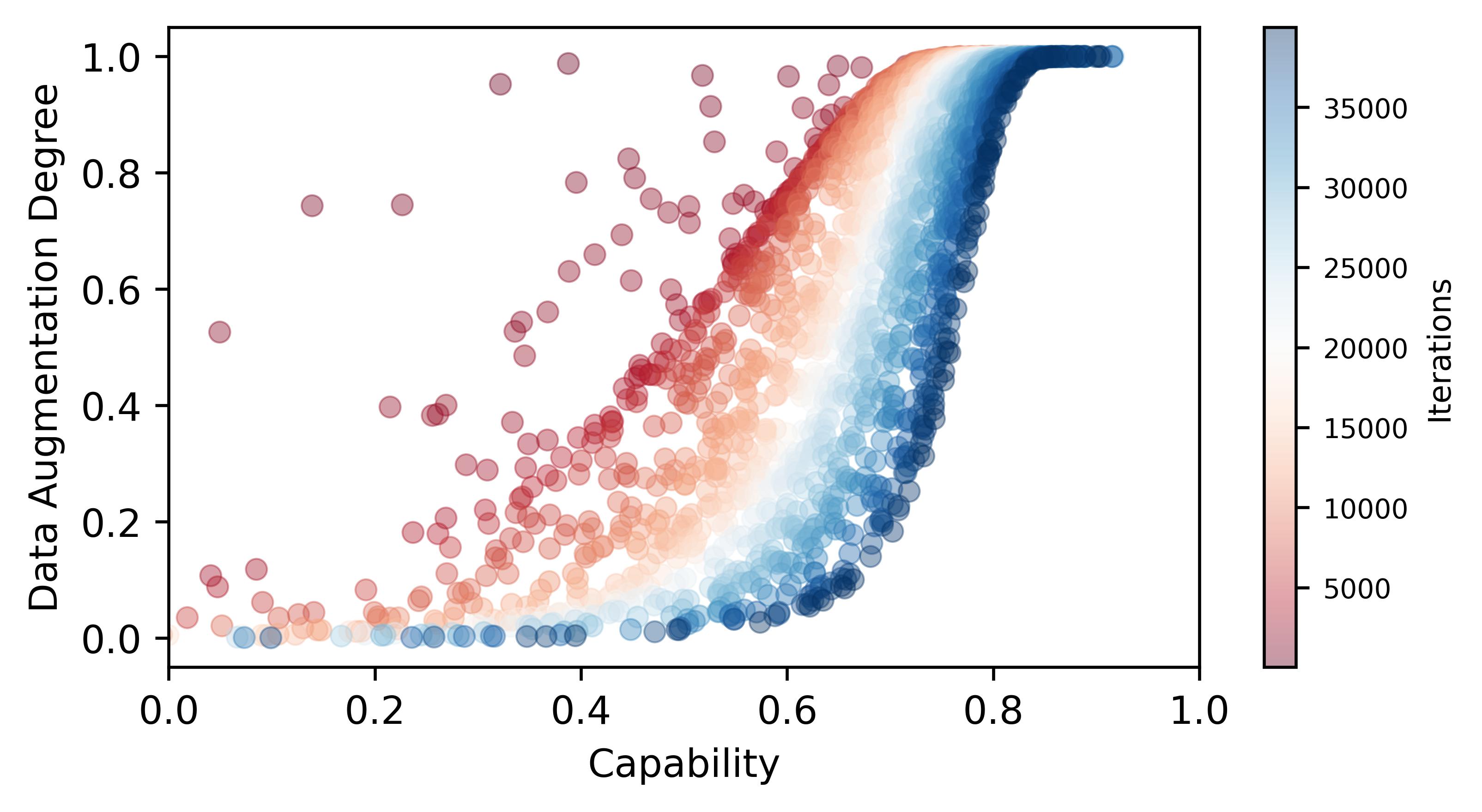}
	\caption{Visualization of the model's capability versus the augmentation degree for new training samples (indicating the difficult level of augmented training samples) along the training iterations. Colors indicate different training iterations, ranging from red to blue as the number of iterations increases. The illustration shows that a low (or high) data augmentation degree is automatically adopted to generate training samples of low (or high) difficult levels at the early (or late) training stage when the capability of contemporarily trained models is low (or high).
 }
	\label{fig:model_skill_vs_difficulty}
\end{figure}

\textbf{Remark 2.} \textit{
The MoDify framework dynamically adjusts the data augmentation degree of training samples in line with the model's capability. Model capability is gauged by the loss of each iteration: 
$M_c = 1.0-\frac{\mathcal{L}_{i} -\mathcal{L}_{min}}{\mathcal{L}_{max} -\mathcal{L}_{min}} $, where $\mathcal{L}_{max}$ and $\mathcal{L}_{min}$ denote the max and min losses in training. Fig.~\ref{fig:model_skill_vs_difficulty} provides an in-depth explanation.
During training, it is noticeable that points with a specific color ranging from red to blue are distributed from the left-bottom corner to the top-right corner, which matches the distribution of the flow channel in Fig.~\ref{fig:flow_theory}. This phenomenon indicates that with the proposed MoDify strategy, the difficulty level of augmented samples increases along with the improvement of the model's capability.
}


MoDify achieves a balance between the model's capability and training samples' difficulty in an online manner, using only an additional forward pass. This approach improves the model's generalization performance by alleviating over-fitting and under-fitting issues.

\subsection{Loss Functions \label{subsec:loss}}

There are two tasks chosen as instantiation to evaluate the effectiveness of the proposed method, including semantic segmentation and object detection. For the semantic segmentation task, the loss function used to supervise between the predicted segmentation result and the ground truth is the Cross Entropy loss~\cite{yi2004automated}. For the object detection task, two losses are adopted, including the \textit{bounding box loss} and the \textit{classification loss}. Specifically, the \textit{bounding box loss} $\mathcal{L}_{\text {bbox}}$ is the smooth L1 loss~\cite{girshick2015fast}, and the \textit{classification loss} $\mathcal{L}_{\text {cls}}$ is the Cross Entropy loss~\cite{yi2004automated}.




\section{Experiments}
This section presents experiments including datasets, metrics, and implementation details, domain generalization evaluations for semantic segmentation and object detection tasks, ablation studies, and discussions respectively. More details are described in the ensuing subsections.


\subsection{Datasets and Metrics}

\textbf{Datasets.}
We evaluate MoDify over multiple datasets across different visual DG tasks on semantic segmentation and object detection, which involve two synthetic source datasets including GTAV~\cite{richter2016playing} and SYNTHIA~\cite{ros2016synthia} and three real target datasets including Cityscapes~\cite{cordts2016cityscapes}, BDD100K~\cite{yu2020bdd100k}, and Mapillary~\cite{neuhold2017mapillary}). GTAV is a large-scale dataset containing 24,966 high-resolution synthetic images with a size of 1914×1052, which shares 19 classes with Cityscapes, BDD100K, and Mapillary. SYNTHIA consists of photo-realistic synthetic images containing 9,400 samples with a resolution of 960×720, which shares 16 classes with the three target datasets. Cityscapes, BDD100K, and Mapillary consist of 2975, 7000, and 18000 real-world training images and 500, 1000, and 2000 validation images respectively.



\textbf{DG for Semantic Segmentation.} We study two synthetic-to-real semantic segmentation tasks, including GTAV $\rightarrow$ \{Cityscapes, BDD100K, Mapillary\} and SYNTHIA $\rightarrow$ \{Cityscapes, BDD100K, Mapillary\}.

\textbf{DG for Object Detection.} We evaluate our methods on several DG scenarios for object detection: SYNTHIA $\rightarrow$ \{ Cityscapes, BDD100K, Mapillary\}.

\textbf{Metrics.} The evaluation metric is the mean Intersection-over-Union (mIoU) for the semantic segmentation task and is the mean Average Precision (mAP) with an IoU threshold equals to 0.5 for the object detection task.

\subsection{Implementation Details}

\textbf{Semantic Segmentation.} We employ DeepLab-V2~\cite{chen2017deeplab} as the segmentation model.  Two backbones are used for experiments, including ResNet-50 and ResNet-101~\cite{he2016deep}. We use SGD~\cite{bottou2010large} with momentum 0.9 as the optimizer. The weight decay is set to $5e^{-4}$ and the learning rate is $2.5e^{-4}$, which is decayed by the polynomial policy~\cite{chen2017deeplab}.

\textbf{Object Detection.} Faster R-CNN~\cite{ren2015faster} is adopted as the detection model. ResNet-101 is used as the backbone. SGD~\cite{bottou2010large} with momentum 0.9 and weight decay $1e^{-4}$ is adopted. The initial learning rate is set to $2e^{-2}$, which is decayed to $2e^{-3}$ and $2e^{-4}$ at the 16 and 22 epochs, respectively.

\begin{figure*}[htbp]
	\centering
	\includegraphics[width=1\linewidth]{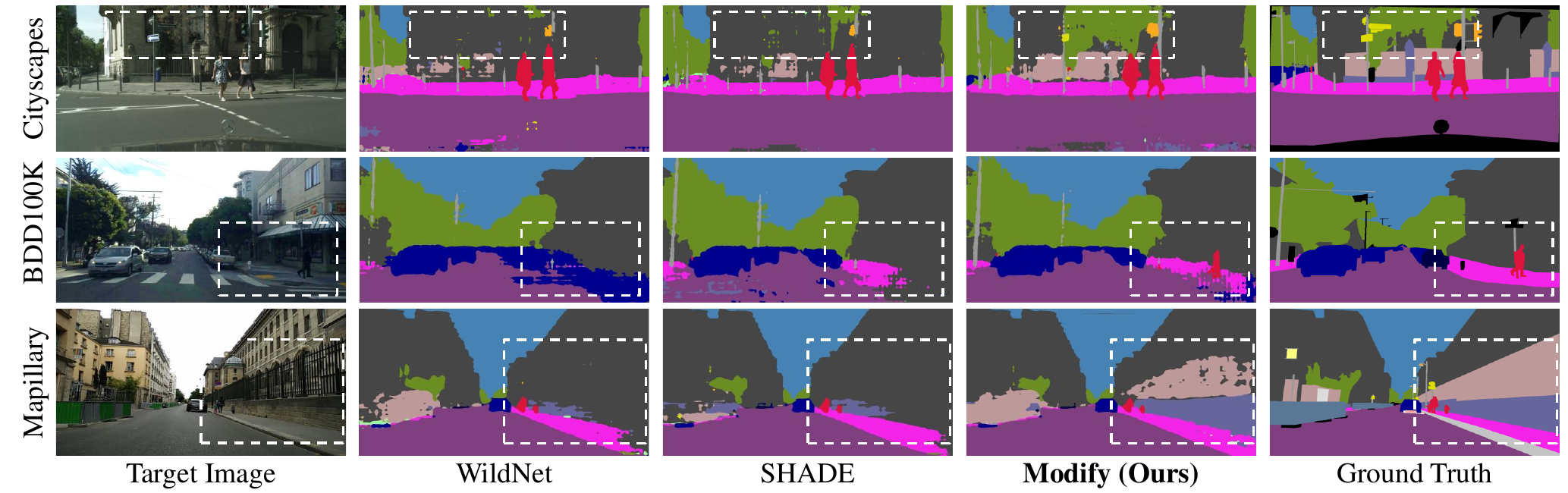}

	\caption{Qualitative illustration of domain generalizable semantic segmentation for GTAV to Cityscapes (Row 1), BDD (Row 2), and Mapillary (Row 3). White boxes highlight regions with clear differences across the compared methods. Compared with other methods, MoDify predicts better building shapes in Row 1, better sidewalk in Row 2, and more accurate fence structures in Row 3.}
	\label{fig:qualitative_result_sem_seg_GTAV}
\end{figure*}

\subsection{Domain Generalizable Semantic Segmentation}

We compare MoDify against state-of-the-art DG-based semantic segmentation methods, including IBN-NET~\cite{pan2018two}, DRPC~\cite{yue2019domain}, GLTR~\cite{peng2021global}, FSDR~\cite{huang2021fsdr}, WildNet~\cite{lee2022wildnet}, and SHADE~\cite{zhao2022style}, on Cityscapes, BDD100K and Mapillary validation sets. The results are reported in Tab.~\ref{tab:sota_comparison_segmentation_GTAV} and Tab.~\ref{tab:sota_comparison_segmentation_synthia} using GTAV and SYNTHIA as source domains respectively. Moreover, we compare the methods using two backbones for a fair comparison. The performance is analyzed in detail as follows:

\begin{table}[]
\centering
\scalebox{0.78}{
\begin{tabular}{c|c|ccc|c}
\hline
Net                         & Method                    & Cityscapes    & BDD100K       & Mapillary     & Mean          \\ \hline\hline
\multirow{7}{*}{\rotatebox{90}{ResNet-101}} & IBN-Net~\cite{pan2018two}          & 37.4          & 34.2          & 36.8          & 36.1          \\
                            & DRPC~\cite{yue2019domain}          & 42.5          & 38.7          & 38.1          & 39.8          \\
                            & GLTR~\cite{peng2021global}         & 43.7          & 39.6          & 39.1          & 40.8          \\
                            & FSDR~\cite{huang2021fsdr}          & 44.8          & 41.2          & 43.4          & 43.1          \\
                            & WildNet~\cite{lee2022wildnet}      & 45.8          & 41.7          & \underline{47.1}          & 44.9          \\
                            & SHADE~\cite{zhao2022style}         & \underline{46.7}          & \underline{43.7}          & 45.5          & \underline{45.3}          \\ 

                            & \textbf{MoDify   (Ours)} \cellcolor[HTML]{EFEFEF}       & \textbf{48.8} \cellcolor[HTML]{EFEFEF}& \textbf{44.2} \cellcolor[HTML]{EFEFEF}& \textbf{47.5}\cellcolor[HTML]{EFEFEF} & \textbf{46.8}\cellcolor[HTML]{EFEFEF} \\
\hline\hline
\multirow{11}{*}{\rotatebox{90}{ResNet-50}} & SW~\cite{pan2019switchable}        & 29.9          & 27.5          & 29.7          & 29.0          \\
                            & IterNorm~\cite{huang2019iterative} & 31.8          & 32.7          & 33.9          & 32.8          \\
                            & ASG~\cite{chen2020automated}       & 31.9          & N/A           & N/A           & N/A           \\
                            & IBN-Net~\cite{pan2018two}          & 33.9          & 32.3          & 37.8          & 34.6          \\
                            & DRPC~\cite{yue2019domain}          & 37.4          & 32.1          & 34.1          & 34.6          \\
                            & ISW~\cite{choi2021robustnet} & 36.6          & 35.2          & 40.3          & 37.4          \\
                            & GLTR~\cite{peng2021global}         & 38.6          & N/A           & N/A           & N/A           \\

                            & SiamDoGe~\cite{wu2022siamdoge}     & 43.0          & 37.5          & 40.6          & 40.4          \\
                            & SHADE~\cite{zhao2022style}         & \underline{44.7}          & \underline{39.3}          & 43.3          & 42.4          \\
                            & WildNet~\cite{lee2022wildnet}      & 44.6          & 38.4          & \underline{46.1}          & \underline{43.0}          \\

                            & \textbf{MoDify   (Ours)} \cellcolor[HTML]{EFEFEF}           &       \textbf{45.7}   \cellcolor[HTML]{EFEFEF}     &   \textbf{40.1} \cellcolor[HTML]{EFEFEF}            &   \textbf{46.2}   \cellcolor[HTML]{EFEFEF}         &      \textbf{44.0}  \cellcolor[HTML]{EFEFEF}       \\
                            \hline

\end{tabular}}
\caption{
    Benchmarking Domain generalization over semantic segmentation task GTAV $\rightarrow$ \{Cityscapes, BDD100K, Mapillary\} in mIoU. Best in \textbf{bold}, second \underline{underlined}.
    }
\label{tab:sota_comparison_segmentation_GTAV}
\end{table}

\textbf{GTAV $\rightarrow$ \{Cityscapes, BDD100K, Mapillary\}.} As shown in Tab.~\ref{tab:sota_comparison_segmentation_GTAV}, the setting GTAV $\rightarrow$ \{Cityscapes, BDD100K, Mapillary\} is used for comparison. MoDify achieves the best performance with 46.8\% and 44.2\% mean mIoU on both backbones. Specifically, when using ResNet-101 as the backbone, MoDify significantly outperforms existing best methods by 2.1\%, 0.5\%, and 2.0\% mIoU on the Cityscapes, BDD100K, and Mapillary datasets, respectively. Moreover, MoDify achieves better performance with ResNet-50, surpassing the second-best methods with 1.8\%, 1.7\%, and 0.1\% mIoU on three datasets, respectively.

\textbf{SYNTHIA $\rightarrow$ \{Cityscapes, BDD100K, Mapillary\}.} Tab.~\ref{tab:sota_comparison_segmentation_synthia} shows the results under the setting SYNTHIA $\rightarrow$ \{Cityscapes, BDD100K, Mapillary\}. We can see that MoDify achieves the best performance on ResNet-50 and ResNet-101 backbones. Specifically, MoDify (ResNet-50) outperforms previous methods with 3.2\%, 2.2\%, and 3.5\% mIoU on the three datasets, respectively. Moreover, MoDify (ResNet-101) improves 2.6\%, 2.1\%, and 2.7\% mIoU as compared with the second-best performance.


As the two tables show, MoDify outperforms
all state-of-the-art DG methods clearly and consistently across both tasks and two network backbones. The superior segmentation performance is largely attributed to the proposed MoDify which balances the model’s capability and the samples’ difficulties along the training process, mitigating the misalignment issue effectively. Moreover, qualitative illustrations in Fig.~\ref{fig:qualitative_result_sem_seg_GTAV} demonstrate the effectiveness of the proposed MoDify which produces better semantic segmentation consistently across different target datasets and domains.

\begin{figure*}[t]
	\centering
	\includegraphics[width=1.0\linewidth]{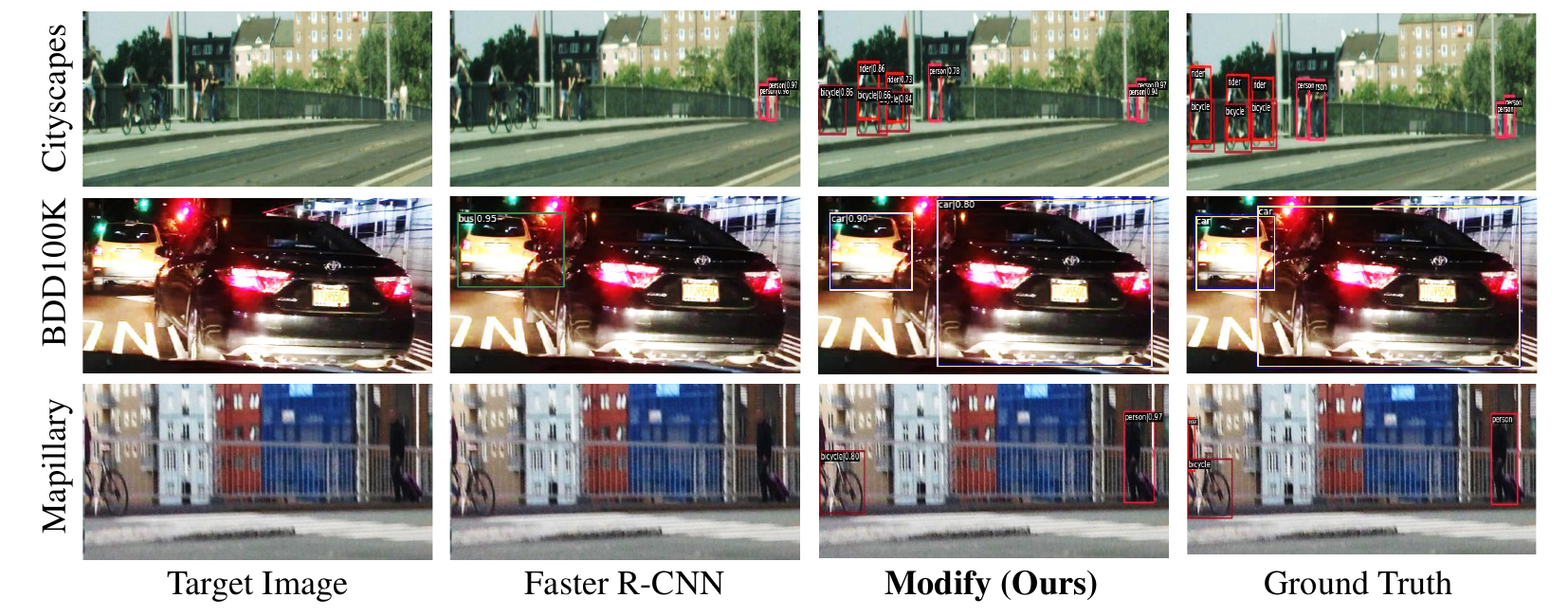}

	\caption{Qualitative illustration of domain generalizable object detection for GTAV to Cityscapes (1st row), BDD (2nd row), and Mapillary (3rd row). The images are cropped and zoomed in for better visualization. In the first row, MoDify performs the best for finding the persons on the sidewalk. In the second row, MoDify predicts both objects accurately with correct categories. In the last row, two objects including a person and a bicycle are predicted correctly by MoDify while Faster R-CNN misses both objects.}
	\label{fig:qualitative_result_object_det}
\end{figure*}

\begin{table}[]
\centering
\scalebox{0.78}{
\begin{tabular}{c|c|ccc|c}
\hline
Net                                         & Method          & Cityscapes & BDD100K & Mapillary & Mean \\ \hline\hline
\multirow{5}{*}{\rotatebox{90}{R101}} & IBN-Net~\cite{pan2018two}  & 37.5       & 33.0    & 33.7      & 34.7 \\
                                            & DRPC~\cite{yue2019domain}  & 37.6       & 34.3    & 34.1      & 35.4 \\
                                            & GLTR~\cite{peng2021global} & 39.7       & 35.3    & 36.4      & 37.1 \\
                                            & FSDR~\cite{huang2021fsdr}  & \underline{40.8}       & \underline{37.4}    & \underline{39.6}      & \underline{39.3} \\
                                            & \textbf{MoDify   (Ours)}
                                            \cellcolor[HTML]{EFEFEF}&   \textbf{43.4}   \cellcolor[HTML]{EFEFEF}      &  \textbf{39.5} \cellcolor[HTML]{EFEFEF}      &  \textbf{42.3}    \cellcolor[HTML]{EFEFEF}     &  \textbf{41.7}  \cellcolor[HTML]{EFEFEF}  \\

\hline\hline
\multirow{2}{*}{\rotatebox{90}{R50}}  & DRPC~\cite{yue2019domain}  & \underline{35.7}       & \underline{31.5}    & \underline{32.7}      & \underline{33.3} \\                         & \textbf{MoDify   (Ours)}
                                            \cellcolor[HTML]{EFEFEF} &   \textbf{38.9} \cellcolor[HTML]{EFEFEF}        &   \textbf{33.7}  \cellcolor[HTML]{EFEFEF}    &    \textbf{36.2}     \cellcolor[HTML]{EFEFEF}  &   \textbf{36.3} \cellcolor[HTML]{EFEFEF} \\ \hline
\end{tabular}
}

\caption{
    Benchmarking domain generalization over semantic segmentation task SYNTHIA $\rightarrow$ \{Cityscapes, BDD100K, Mapillary\} in mIoU. R50 and R101 represent ResNet-50 and ResNet-101, respectively. Best in \textbf{bold}, second \underline{underlined}.
    }
\label{tab:sota_comparison_segmentation_synthia}
\end{table}

\begin{table}[]

\centering
\scalebox{0.8}{
\begin{tabular}{c|ccc|c}
\hline
Method                  & Cityscapes    & BDD100K       & Mapillary     & Mean          \\ \hline\hline
Faster R-CNN~\cite{ren2015faster} & 24.3         &  20.1        & 20.8         & 21.7         \\
IBN-Net\cite{pan2018two} & 30.1          & 23.1          & 22.3          & 25.1          \\
FSDR\cite{huang2021fsdr} & \underline{33.5}          & \underline{25.2}          & \underline{24.9}          & \underline{27.8}          \\
\rowcolor[HTML]{EFEFEF}
\textbf{MoDify (Ours)}          & \textbf{37.0} & \textbf{26.1} & \textbf{26.9} & \textbf{30.0} \\ \hline
\end{tabular}
}
\caption{
    Benchmarking domain generalization over object detection task SYNTHIA $\rightarrow$ \{Cityscapes, BDD100K, Mapillary\} in mIoU. Faster-RCNN with ResNet-101 is the base framework for all methods. Best in \textbf{bold}, second \underline{underlined}.
}
\label{tab:sota_comparison_detection_faster_rcnn}
\end{table}


\subsection{Domain Generalizable Object Detection}

Apart from semantic segmentation, the object detection task is also used to evaluate the effectiveness of the proposed method over the DG-based object detection tasks SYNTHIA $\rightarrow$ \{Cityscapes, BDD100K, Mapillary\}. State-of-the-art methods IBN-NET~\cite{pan2018two} and FSDR~\cite{huang2021fsdr} are used to compare with MoDify. As shown in Tab.~\ref{tab:sota_comparison_detection_faster_rcnn}, on all three datasets, MoDify beats the second-best performance by 3.5\%, 0.9\% and 2.0\% mAP improvements, respectively. The qualitative comparison results are shown in Fig.~\ref{fig:qualitative_result_object_det}, showing cases of all three datasets. Compared with the results of other methods, the detection results of MoDify are more precise with fewer false positive predictions.

\subsection{Ablation Study}

We examine different MoDify designs to find out how they contribute to network generalization in semantic segmentation. Specifically, we trained six models over the UDG task GTAV $\rightarrow$ Cityscapes, and Tab.~\ref{tab:ablation_study_modules} presents the corresponding experimental results.

\begin{table}[]
\centering
\scalebox{0.9}{
\begin{tabular}{c|ccc|c}
\hline
Index & RGB Shuffle & MoDify-DA  & MoDify-NO  & mIoU          \\ \hline\hline
1     &                   &            &            & 36.6          \\
2     & \checkmark        &            &            & 43.8          \\
3     & \checkmark        & \checkmark &            & 46.3          \\
4     &                   &            & \checkmark &   43.3            \\
5     & \checkmark        &            & \checkmark &   46.5            \\
\rowcolor[HTML]{EFEFEF}
6     & \checkmark        & \checkmark & \checkmark & \textbf{48.8} \\ \hline
\end{tabular}
}
\caption{
    Ablation study of the proposed components in MoDify over the domain generalizable semantic segmentation task GTAV $\rightarrow$ Cityscapes, using ResNet-101 as the backbone.
    }
\label{tab:ablation_study_modules}
\end{table}

We can observe that the baseline trained with the GTA data only does not perform well due to domain bias. MoDify-DA and MoDify-NO outperform the Baseline by a significant margin, which demonstrates the importance of balancing the seesaw between the model's capability and the samples' difficulty level throughout the training process to achieve domain-generalizable models. When combined with RGB Shuffle, MoDify-NO achieves slightly better results than MoDify-DA, which can be largely attributed to its direct connection to network updates during training. Furthermore, MoDify consistently achieves the best performance. This indicates that MoDify-DA and MoDify-NO are complementary, where the two designs work collaboratively to generate difficulty-aware augmentation samples and coordinate the augmentation and network training smoothly. Besides, the Color shuffle augmentation strategy improves the Baseline by a large margin, demonstrating the effectiveness of this simple yet effective technique.




\subsection{Discussions}
This section covers three main parts, including analysis on the losses of different methods during training, the compatibility of our method with existing approaches, as well as its computational cost.

\begin{figure}[t]
	\centering
        \includegraphics[width=\linewidth]{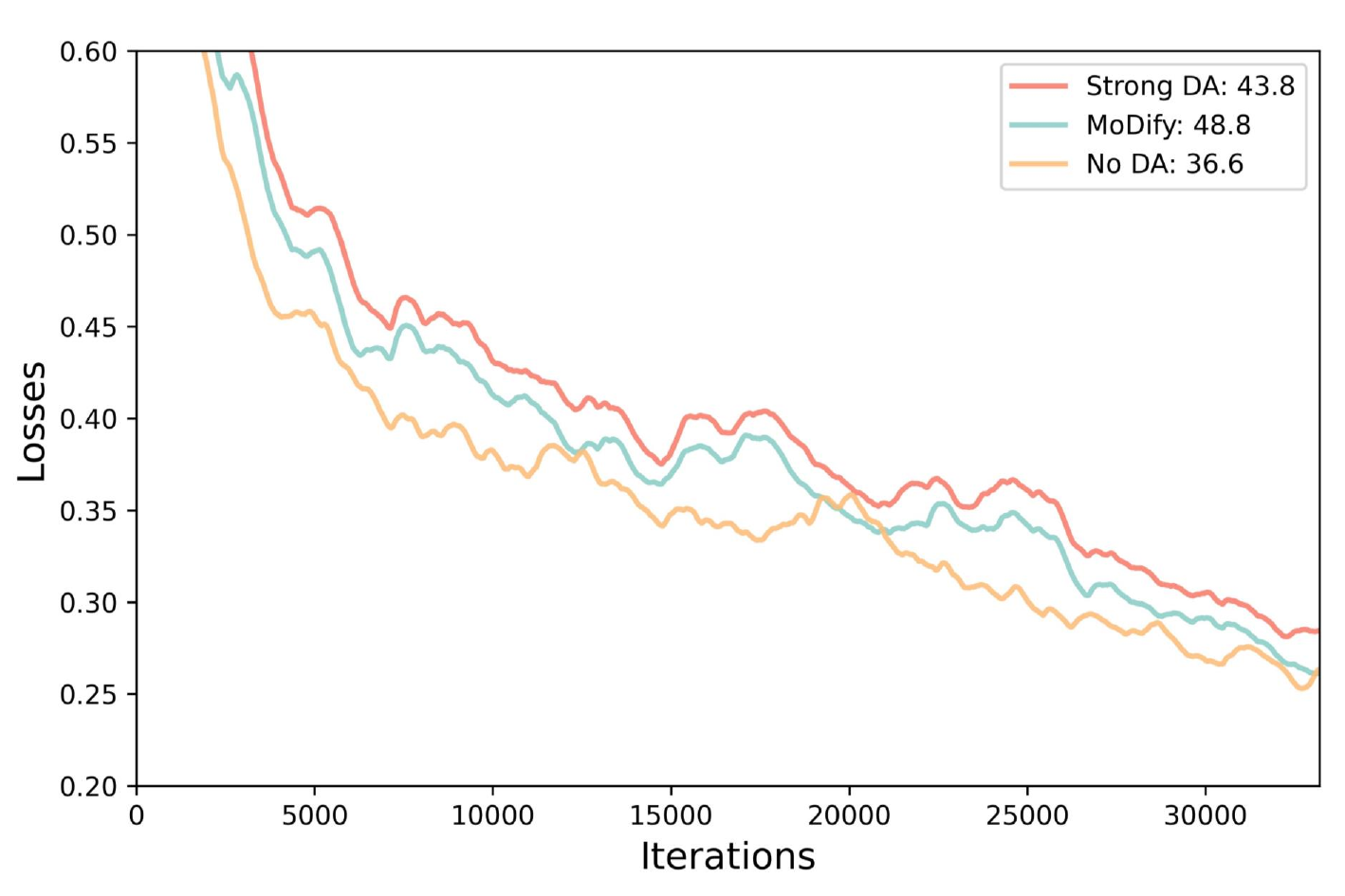}
	\caption{Visualization of the losses during the training process for strong data augmentation (Strong DA), MoDify, and no data augmentation (No DA), respectively. No data augmentation results in low loss but poor performance due to over-fitting. Strong data augmentation leads to high loss and sub-optimal performance due to under-fitting. Our method achieves the best performance with moderate loss, indicating that MoDify alleviates the misfitting issue effectively. Results are obtained on the semantic segmentation task from GTAV~\cite{richter2016playing} to Cityscapes~\cite{cordts2016cityscapes} with ResNet-101.}
	\label{fig:fitting_analysis_v1}
\end{figure}

Fig.~\ref{fig:fitting_analysis_v1} shows the loss curves of the methods using strong data augmentation, no data augmentation, and MoDify during training, as well as the mIoU performance in the target domain.
As shown in Fig~\ref{fig:fitting_analysis_v1}, we can observe that the proposed MoDify balances the model's fitting to the source domain data between the Strong DA-based method and the No DA-based method during training.
The method with no data augmentation has the lowest loss during training, but the worst performance due to over-fitting on the source domain (as illustrated in \textcolor[RGB]{238,173,14}{yellow} line). Strong data augmentation leads to high loss and sub-optimal performance due to under-fitting on the source domain (as illustrated in \textcolor[RGB]{255,0,0}{red} line). Our method achieves the best performance with moderate loss, indicating that MoDify alleviates the misfitting issue during training (as illustrated in \textcolor[RGB]{60,179,113}{green} line).

MoDify is complementary to existing domain generalization networks, which can be easily incorporated into them with consistent performance boosts but little extra parameters and computation. We evaluated this feature by incorporating MoDify's training strategies into two competitive domain generalization networks including ISW~\cite{choi2021robustnet} and SHADE~\cite{zhao2022style} as shown in Tab.~\ref{tab:ablation_study_combine_with_other_method}. During training, we balance the difficulty level of the augmented samples and the capability of networks. As Tab.~\ref{tab:ablation_study_combine_with_other_method} shows, the incorporation of MoDify improves the semantic segmentation of state-of-the-art networks consistently. As the incorporation of MoDify just includes a training-free loss bank without changing network structures, the inference has few extra parameters and computation once the model is trained.

To validate the computational efficiency of our proposed method, we conducted a detailed analysis of its parameters, floating-point operations per second (FLOPs), and inference time. The results are presented in Tab.~\ref{tab:computational_cost}. It can be observed that our approach incurs only a minor additional computational cost during testing. Additionally, we calculated that the training time required by our proposed method is only 1.4 times that of the baseline method's training time.
These findings confirm the practicality and efficiency of our proposed method.


\begin{table}[]
\centering
\scalebox{0.88}{
\begin{tabular}{@{}l|llllll@{}}
\hline
\multicolumn{1}{c|}{}        & \multicolumn{2}{c}{Cityscapes}            & \multicolumn{2}{c}{BDD100K}               & \multicolumn{2}{c}{Mapillary} \\ \cline{2-7}
                             & Base & +Ours                              & Base & +Ours                              & Base      & +Ours             \\ \hline \hline
ISW~\cite{choi2021robustnet} & 36.6 & \multicolumn{1}{l|}{\textbf{40.8}} & 35.2 & \multicolumn{1}{l|}{\textbf{38.0}} & 40.3      & \textbf{43.7}     \\
SHADE~\cite{zhao2022style}   & 44.7 & \multicolumn{1}{l|}{\textbf{46.0}} & 39.3 & \multicolumn{1}{l|}{\textbf{40.5}} & 43.3      & \textbf{44.2}     \\ \hline
\end{tabular}
}
\caption{
    MoDify's training strategies are complementary to existing domain generalization methods. For the task GTAV $\rightarrow$ \{Cityscapes, BDD100K, Mapillary\} (using ResNet-50 as the backbone), including MoDify (Ours) consistently boosts the performance of domain generalization.
    }
\label{tab:ablation_study_combine_with_other_method}
\end{table}

\begin{table}[]
\centering
\scalebox{1.0}{
\begin{tabular}{c|c|c|c}
\hline
Models                        &  Params & FLOPs & Inference Time \\ \hline \hline
ISW~\cite{choi2021robustnet} &   45.1M
          & 556.22G       &     13.6ms                \\
WildNet~\cite{lee2022wildnet} &   45.1M          &  556.19G      &  21.3ms                   \\
SHADE~\cite{zhao2022style}    &   45.1M         &  556.19G        &  14.5ms                   \\
\rowcolor[HTML]{EFEFEF}
Ours                          &   45.1M           &  556.19G      &   13.7ms                  \\
\hline
\end{tabular}
}
\caption{
   Comparison of computational cost. The tests are carried out using the image size of 1024 $\times$ 2048 on NVIDIA V100 GPU.
}
\label{tab:computational_cost}
\end{table}

\section{Methodology Limitation}
MoDify enhances the model's generalization ability by adjusting the strength of image-level data augmentation during the training process, which leads to better performance of visual task models in several challenging scenarios. However, there are certain region-level difficult samples where the performance of the model is relatively poor. To illustrate this, some failure cases are visualized in the supplementary material. 
Future work could explore a fine-grained region adaptive strategy, applying data augmentation with appropriate levels to different image regions, which is a more targeted approach.

\section{Conclusion}
This paper presents the Momentum Difficulty (MoDify) technique that tackles domain generalization challenges by mitigating the misalignment between the overall difficulty degree of training samples and the capability of the contemporary deep network model along with the training process. Specifically, we designed MoDify-based Data Augmentation (MoDify-DA) and MoDify-based Network Optimization (MoDify-NO), which coordinate the augmentation and the network training smoothly. The proposed MoDify has three valuable features: 1) it is generic to various visual recognition tasks with consistently superior performance; 2) it is an online yet lightweight technique in various downstream; 3) it complements with existing domain generalization methods with consistent performance boosts. 

\section*{Acknowledgement}
This study is supported under the RIE2020 Industry Alignment Fund – Industry Collaboration Projects (IAF-ICP) Funding Initiative, as well as cash and in-kind contribution from the industry partner(s).


{\small
\bibliographystyle{ieee_fullname}
\bibliography{egbib}
}

\end{document}